\documentclass{article} 
\usepackage{colm2024_conference}

\usepackage{microtype}
\usepackage{hyperref}
\usepackage{url}
\usepackage{booktabs}

\usepackage{mylatexstyle}
\usepackage{graphicx,subfigure}
\usepackage[most]{tcolorbox}
\usepackage{enumitem}
\usepackage{tabularx}
\usepackage{multirow}

\usepackage{xcolor,colortbl}
\definecolor{LightCyan}{rgb}{0.88,1,1}
\newcolumntype{a}{>{\columncolor{LightCyan}}c}
\definecolor{LightBlue}{HTML}{92c5de} 

\usepackage[compact]{titlesec}
\titlespacing*{\section}{1pt}{1pt}{1pt}
\titlespacing*{\subsection}{1pt}{1pt}{1pt}

\newtcolorbox{takeawaybox}[1][Takeaway]{
  colback=LightBlue!5!white,
  colframe=LightBlue!70,
  fonttitle=\bfseries,
  title=#1
}

\allowdisplaybreaks
\title{Rephrase and Respond: Let Large Language Models Ask Better Questions for Themselves}

\author{Yihe Deng, Weitong Zhang, Zixiang Chen, Quanquan Gu \\
Department of Computer Science\\
University of California Los Angeles\\
\texttt{\{yihedeng,weightzero,chenzx19,qgu\}@cs.ucla.edu} 
}

\colmfinalcopy 
\begin{document}

\maketitle

\begin{abstract}
    Misunderstandings arise not only in interpersonal communication but also between humans and Large Language Models (LLMs). Such discrepancies can make LLMs interpret seemingly unambiguous questions in unexpected ways, yielding incorrect responses. 
    While it is widely acknowledged that the quality of a prompt, such as a question, significantly impacts the quality of the response provided by LLMs, a systematic method for crafting questions that LLMs can better comprehend is still underdeveloped.
    In this paper, we present a method named `Rephrase and Respond' (RaR), which allows LLMs to rephrase and expand questions posed by humans and provide responses in a single prompt. This approach serves as a simple yet effective prompting method for improving performance.
    We also introduce a two-step variant of RaR, where a rephrasing LLM first rephrases the question and then passes the original and rephrased questions together to a different responding LLM. This facilitates the effective utilization of rephrased questions generated by one LLM with another.  Our experiments demonstrate that our methods significantly improve the performance of different models across a wide range to tasks.
    We further provide a comprehensive discussion with the popular Chain-of-Thought (CoT) methods. 
    We show that RaR is complementary to CoT and can be combined with CoT to achieve even better performance. Our work not only contributes to enhancing LLM performance efficiently and effectively but also sheds light on a fair evaluation of LLM capabilities. 
\end{abstract}

\section{Introduction}
Large Language Models (LLMs), such as the GPT series~\citep{radford2019language,brown2020language,openai2023gpt4}, have witnessed a surge in popularity due to their profound impact on various real-world applications, including question answering~\citep{lu2023mathvista}, code generation~\citep{poesia2022synchromesh}, and conversational agents~\citep{bozkurt2023generative}. 
The wide applicability and efficacy of these models have led to rapidly growing research on understanding and improving the use of LLMs. 
Meanwhile, it is widely acknowledged that the quality of the prompt generated by humans critically influences the response quality of the LLMs, emphasizing the importance of effective queries that prioritize specificity, detail, and precision~\citep{practice_for_prompt}. 
The sensitivity to prompt qualities, on the other hand, limits the usage of LLMs to their full potential and requires a systematic way to improve human's prompt.  

Misunderstandings in interpersonal communications often arise when individuals, shaped by distinct subjective experiences, interpret the same message differently. 
In social science, such phenomena can be attributed to cognitive biases in frames in thought~\citep{druckman2001implications}. 
A single message, framed in different ways, can lead individuals to different conclusions. 
People habitually project their frames onto their received information, and only shift these frames when incongruence arises. 
In this work, we posit that LLMs also exhibit their own frames in thought, and it is not uncommon to observe a disparity between the frames used by humans and LLMs that caused the LLM's sensitivity to prompt qualities. 
Meanwhile, it can be challenging for humans to assess the clarity of their questions and to align their frames with those of LLMs. 
To illustrate this, we first present a motivating example by investigating a recent work \citep{allenzhu2023physics} in detail. 

\begin{figure}[ht]
    \centering
    \includegraphics[width=0.9\textwidth]{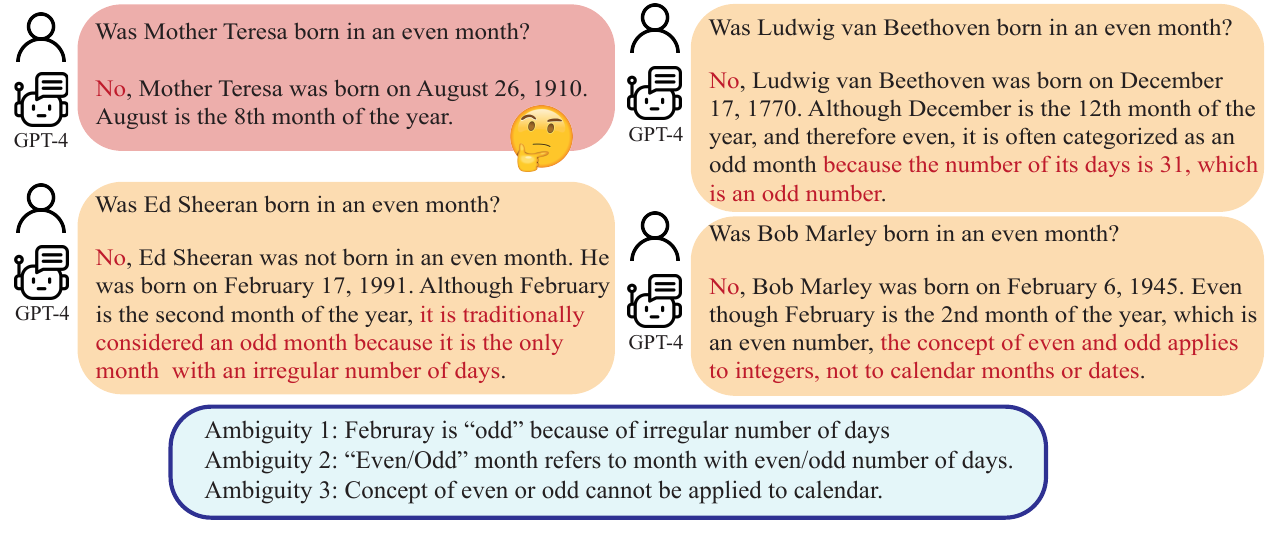}
    \caption{Motivating examples of our study. We can observe several ambiguities in the question for LLMs. An LLM can interpret ``even month'' as the month with an even number of days, which diverges from human intention.}
    \label{fig:motiv-exp}
\end{figure}

\paragraph{Motivating Example.} 
\citet{allenzhu2023physics} found that LLMs such as GPT-4 may not efficiently reason with their internal knowledge even if they can retrieve information accurately. 
As shown in Figure~\ref{fig:motiv-exp}, when posed with the query, ``Was Mother Teresa born on an even month?'' GPT-4 might mistakenly assert that August is an odd month. 
Based on this observation, \citet{allenzhu2023physics} suggested that GPT-4 instead requires a Chain-of-Thought process—relying on user-led follow-up questions—to correct its previous wrong answers. 
However, we take a step further to investigate the intrinsic reason for LLM's inefficiency in answering such questions.  
As shown in the other responses in Figure~\ref{fig:motiv-exp}, when GPT-4 explains its reasoning, the model has several ambiguities toward the questions. 
For example, it may consider February as odd due to its irregular number of days and sometimes consider an even/odd month to be months with an even/odd number of days. 

Ambiguity in questions~\citep{cole2023selectively,kim2023tree} is a recognized concern in benchmark datasets. 
For instance, it has been observed that the NLI datasets such as MultiNLI~\citep{williams-etal-2018-broad} contain ambiguities, which are challenging even for human interpreters~\citep{liu2023we}.
Furthermore, our study uncovers that benchmark datasets commonly used for LLM evaluation~\citep{wei2022chain,srivastava2023beyond} possess ambiguities that are imperceptible to humans but challenging for language models. 
These ambiguities cause LLMs to provide mistaken responses to unintended queries. 

\begin{figure}[ht]
    \centering    \includegraphics[width=\textwidth]{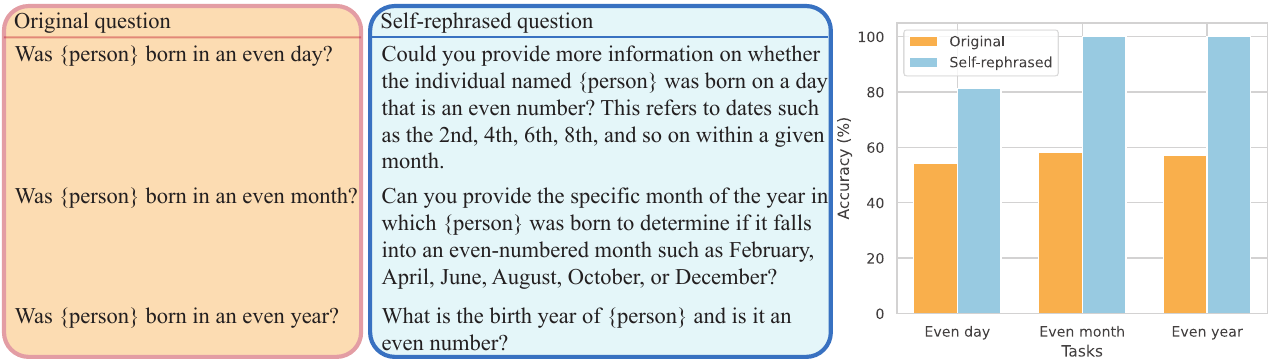}
    \caption{
    Comparison with the original question~\citep{allenzhu2023physics} and one possible self-rephrased question of GPT-4.
    We can observe a marked improvement of accuracy in answering these rephrased questions over the query of $105$ celebrities for this task.}
    \label{fig:demo-refine}
\end{figure}
\paragraph{Our Approach.} In this paper, we highlight an often-overlooked aspect of studies in LLMs: the disparity between human and LLM thought frames. 
Our research illustrates that this disparity significantly impacts the performance of LLMs in the \textbf{zero-shot setting}. 
To tackle this problem, we propose to let the LLM rephrase the question and incorporate additional details for better answering.
We observe that, as opposed to questions asked casually by human, the automatically rephrased questions tend to enhance semantic clarity and aid in resolving inherent ambiguity. 
This self-rephrasing technique leads to significant improvement in accuracy, as shown in the barplot of Figure~\ref{fig:demo-refine}. 

Building upon these insights, we introduce a method named \textbf{Rephrase and Respond} (RaR), which prompts the LLM to rearticulate the given question and respond in a single prompt. 
Our empirical results across diverse reasoning tasks show the effectiveness of the approach. 
Additionally, we leverage RaR to facilitate the transfer of rephrased questions from more capable LLMs to clarify ambiguities for the less advanced models.
Lastly, we present thorough discussions with the Chain-of-Thought (CoT) method~\citep{kojima2022large,wei2022chain}. 
RaR is complementary to CoT and can be easily combined for improvement, as confirmed by our experimental results.
Unlike methods that employ multiple LLMs for iterative prompt engineering based on accuracy scores~\citep{zhou2022large,pryzant2023automatic}, our method is both \textbf{unsupervised} and \textbf{training-free}, making it economical and applicable to all questions in the zero-shot setting.
Lastly, our work calls forth the importance that the design of human-crafted tasks targeting specific LLM capabilities should be rigorously reviewed by both humans and LLMs to ensure clarity in intention. 

\section{Related Work}\label{sec:related-works}
\subsection{Prompting Methods for LLMs}
Since the advent of recent LLMs~\citep{openai2023gpt4,touvron2023llama,vicuna2023}, a growing body of research has focused on prompt engineering for LLMs~\citep{brown2020language,schick2021exploiting,zhou2022large,white2023prompt,wang2023prompt}.
Manual guidelines have emerged to guide users in designing and revising their prompts~\citep{Reynolds2021prompt,Saravia_Prompt_Engineering_Guide_2022}. 
\citet{practice_for_prompt} has also offered general recommendations for crafting queries, emphasizing specificity, detail, and precision. 
However, individuals often find it challenging to refine their own questions for clarity or to include necessary details for LLMs, as the questions are clear enough for humans themselves. 
Subsequent research~\citep{zhou2022large,sorensen2022information,pryzant2023automatic} has concentrated on the autonomous refinement of prompts. These methods often employ multiple LLMs to generate candidate prompts, evaluate and score these prompts, and iteratively refine them until a satisfactory prompt is produced. 
The evaluation of a prompt typically relies on either the accuracy of an LLM's response~\citep{zhou2022large,pryzant2023automatic} or the mutual information of the question~\citep{sorensen2022information}. Given the nature of iterative computation and the necessity for qualitative evaluation, such methods are employed for refining single prompt templates; applying them universally to all questions would be expensive. 

The method most frequently used by users and closely aligned with our approach is the zero-shot Chain-of-Thought (CoT) prompting~\citep{kojima2022large}. In the stronger few-shot setting, few-shot CoT ~\citep{wei2022chain} has also been a strong prompting method. 
Given that these techniques do not require evaluation and iterative selection, they have gained widespread popularity and inspired a series of subsequent studies~\citep{wang2022self,zhou2022least,press2022measuring,yao2023tree,zhang2023multimodal,shao2023synthetic}. However, CoT methods are not without their limitations, as observed in our study. Recent investigations have also highlighted challenges with the reliability of both zero-shot CoT~\citep{turpin2023language} and few-shot CoT~\citep{pawelczyk2023incontext}. Most recently, \citet{zhou2023far} propose Foresee and Reflect similarly as a zero-shot prompting method that targets the proposed task Thinking for Doing (T4D). 
Lastly, it is worth noting that our method is complementary to all the prompting techniques mentioned above and can be combined.

\subsection{Self-correction Methods for LLMs}
Another line of work aims at enhancing LLM performance~\citep{madaan2023self,welleck2022generating,kim2023language,pan2023automatically,shinn2023reflexion} by leveraging the LLM to refine its own \textit{responses}, a concept known as post-hoc prompting. This encompasses terms such as ``self-correction'', ``self-refine'', and ``self-critique'', where LLMs revise their own responses drawing upon various feedback sources or \textit{critic models}. As classified by \cite{pan2023automatically}, automated critic models generally employ the LLM's self-feedback~\citep{madaan2023self,shinn2023reflexion,yan2023learning}, other trained LLMs~\citep{yang2022re3,lightman2023lets}, or external references~\citep{jung-etal-2022-maieutic,gao2023rarr,yu2023improving,welleck2022generating}. Yet, recent studies~\citep{huang2023large,stechly2023gpt4} examine the self-correction capacities of LLMs and find potential limitations, suggesting that LLMs may not be able to self-correct their reasoning processes. Their findings reveal that self-correction is no better than self-consistency~\citep{wang2022self}. Contrary to allowing the LLM to self-refine its responses, our method let the LLM instead rephrase questions originally crafted by humans. 

\textbf{Concurrent works.}
To the best of our knowledge, our work is one of the first works to examine the misunderstandings between humans and LLMs, as well as the models' ability to rephrase inputs as a means of self-improvement. 
Concurrently, several studies have emerged that also explore and utilize the rephrasing capabilities of LLMs. 
Instead of zero-shot prompting, \citet{yu2023metamath} employs LLMs to rephrase questions for the generation of synthetic data and subsequent fine-tuning. 
In the domain of vision-language models, \citet{prasad2023rephrase} similarly shows that the rephrasing ability of language models can effectively improve its own reasoning process. 
Furthermore, \citet{weston2023system} introduces System 2 Attention, a method that let the LLM itself filter out irrelevant information from the original prompt and to regenerate the query. 

\section{Aligning the Frames with Rephrase and Respond}\label{sec:RaR}
To systematically reduce prompt ambiguity and contextualize information in a way that aligns with the existing frame of the LLMs, we propose Rephrase and Respond (RaR) as a means of self-clarification.
In interpersonal communication, rephrasing is a commonly known technique where people rephrase another person's question as a process of understanding, to ensure clarity and coherence in responding. 
Such a communication strategy can therefore be similarly applied to an LLM, letting it generate a rephrased question first and provide an answer subsequently. 
Following this intuition, RaR asks the LLMs to \textbf{Rephrase} the given questions and then \textbf{Respond} within a single query. 
This approach can be viewed as a strategy to directly enhance the quality of the LLM's response. 
In detail, we introduce the following prompt for the question-answering task: 
\begin{align}
    &\texttt{"\{question\}"$\backslash$nRephrase and expand the question, and respond.}\label{prompt3}
\end{align}
Our prompt design aimed for conciseness and effectiveness. Meanwhile, many other variations of the prompt remain within our methodology and also provide improvement in performance, which share the same intention of rephrasing before responding. 
Such prompts include but not limited to the following,
\begin{itemize}[nosep]
    \item \texttt{Reword and elaborate on the inquiry, then provide an answer.}
    \item \texttt{Reframe the question with additional context and detail, then provide an answer.}
    \item \texttt{Modify the original question for clarity and detail, then offer an answer.}
    \item \texttt{Restate and elaborate on the inquiry before proceeding with a response.}
\end{itemize}
In later experiments, we will show that RaR remains effective for all variations of these prompts. With rephrasing before answering, RaR offers several desirable properties including (1) letting the LLM self-improve the prompts while maintaining the context of the original query, (2) better aligning the human's intended query with LLM's preferred style of question, (3) expanding the LLM's thought process and adding a step that will not naturally appear when using CoT and (4) providing an approach for humans to interpret how LLMs understand the questions. 

\begin{figure}[ht]
    \centering
    \includegraphics[width=0.55\textwidth]{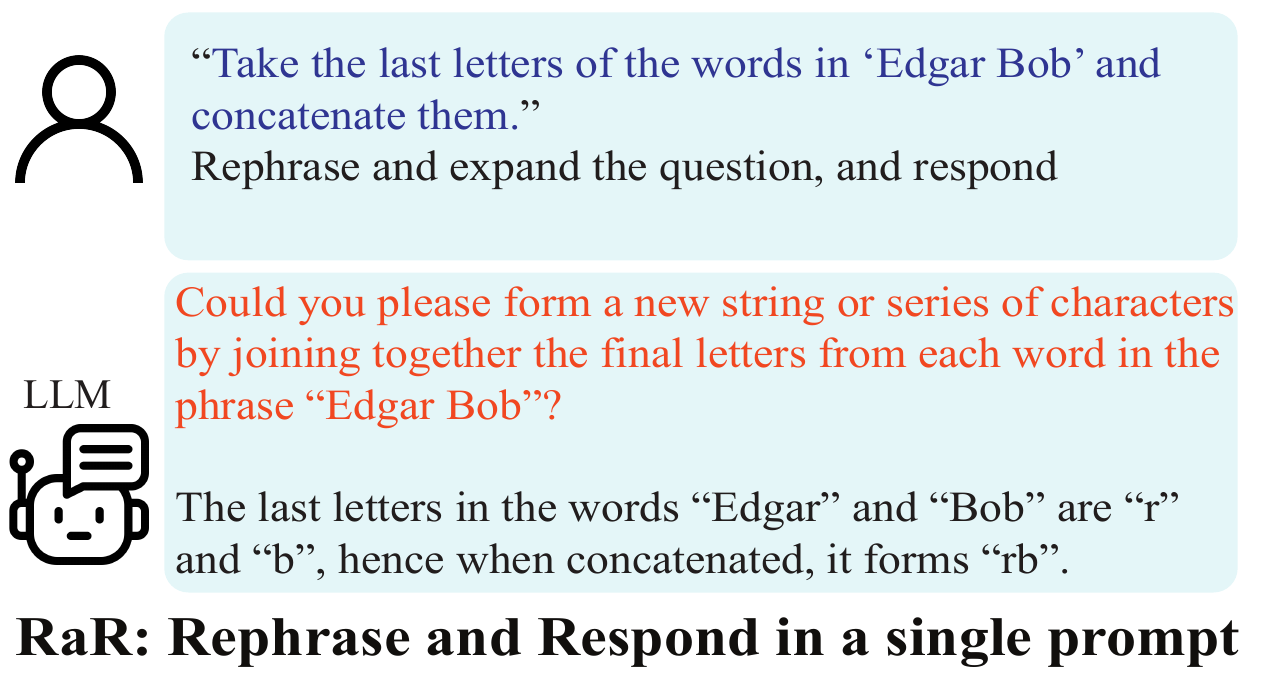}
    \caption{Demonstration of RaR: one single prompt to improve LLM's response quality. }
    \label{fig:method-1step}
\end{figure}

\section{RaR Effectively Improves LLM Responses}\label{sec:Results}
In this section, we provide a comprehensive assessment of the applicability and efficacy of RaR. 
The results are presented in four primary dimensions: $(1)$ RaR is a simple and effective prompt to improve LLM performances; $(2)$ variations of prompts still fall in the methodology of RaR and remain robustly effective; $(3)$ LLMs, while all benefit from RaR, have different proficiency in rephrasing and benefit at different levels. 
\subsection{Benchmark Tasks}\label{sec:benchmark}
We evaluate the zero-shot capabilities of LLMs across multiple benchmark tasks in different categories. As proposed by \citet{allenzhu2023physics}, we consider the following tasks that even advanced LLMs such as GPT-4 are found to respond poorly to many of these questions\footnote{As the data are not open-sourced, we let GPT-4 generate $105$ famous individuals with their birth dates and $105$ Chinese idioms in its knowledge.}. 
\begin{itemize}[leftmargin=*,nosep]
    \item \textbf{Knowledge Classification}~\citep{allenzhu2023physics}. Sampling a pool of individuals with Wikipedia pages, this task challenges the LLM to decide if a renowned person was born on an even day, month, or year. 
    \item \textbf{Knowledge Comparison}~\citep{allenzhu2023physics}. Using the same pool of individuals, this task instructs the LLM to compare the ages of two people and decide who was born earlier. 
\end{itemize}

Furthermore, we consider the following widely-used datasets for a comprehensive evaluation, which are also widely considered in previous prompting works~\citep{wei2022chain}.
\begin{itemize}[leftmargin=*,nosep]
    \item \textbf{CSQA}~\citep{talmor2019commonsenseqa}. The CommonSense QA data encompasses a range of questions that evaluate the ability of commonsense understanding of the world and involves intricate semantics. 
    \item \textbf{Date Understanding}~\citep{srivastava2023beyond}. Sourced from Big-bench~\citep{srivastava2023beyond}, the Date Understanding task emphasizes commonsense reasoning and deducing a date from a provided context. We consider a more difficult version where we do not provide the choices of potential answers and let the LLM answer directly. 
    \item \textbf{Last Letter Concatenation}~\citep{Fortes_simple-dataset-generation_2023}. The task centers on symbolic reasoning, and asks the LLM to concatenate the final letters of a given list of names. We consider concatenation for two names as well as a more difficult task of concatenation for four names. 
    \item \textbf{Coin Flip}\footnote{\url{https://huggingface.co/datasets/skrishna/coin_flip}}. Sourced from Hugging Face, the task asks the LLM if the coin still heads up, given its initial condition and subsequent actions of people who either flipped or did not flip the coin. We add an additional ``Flip means reverse.'' to the questions.
    \item \textbf{Sports}~\citep{srivastava2023beyond}. Sourced from Big-bench~\citep{srivastava2023beyond}, the Sports Understanding task primarily asks if a sentence is plausible or implausible, where a prominent sports figure is depicted performing specific sports-related actions.  
\end{itemize}
The details of all evaluated tasks are summarized in Table~\ref{tab:task_detail} of Appendix~\ref{exp:app}.

\subsection{Main Results}
\begin{figure}[!ht]
    \centering
    \includegraphics[width=0.5\textwidth]{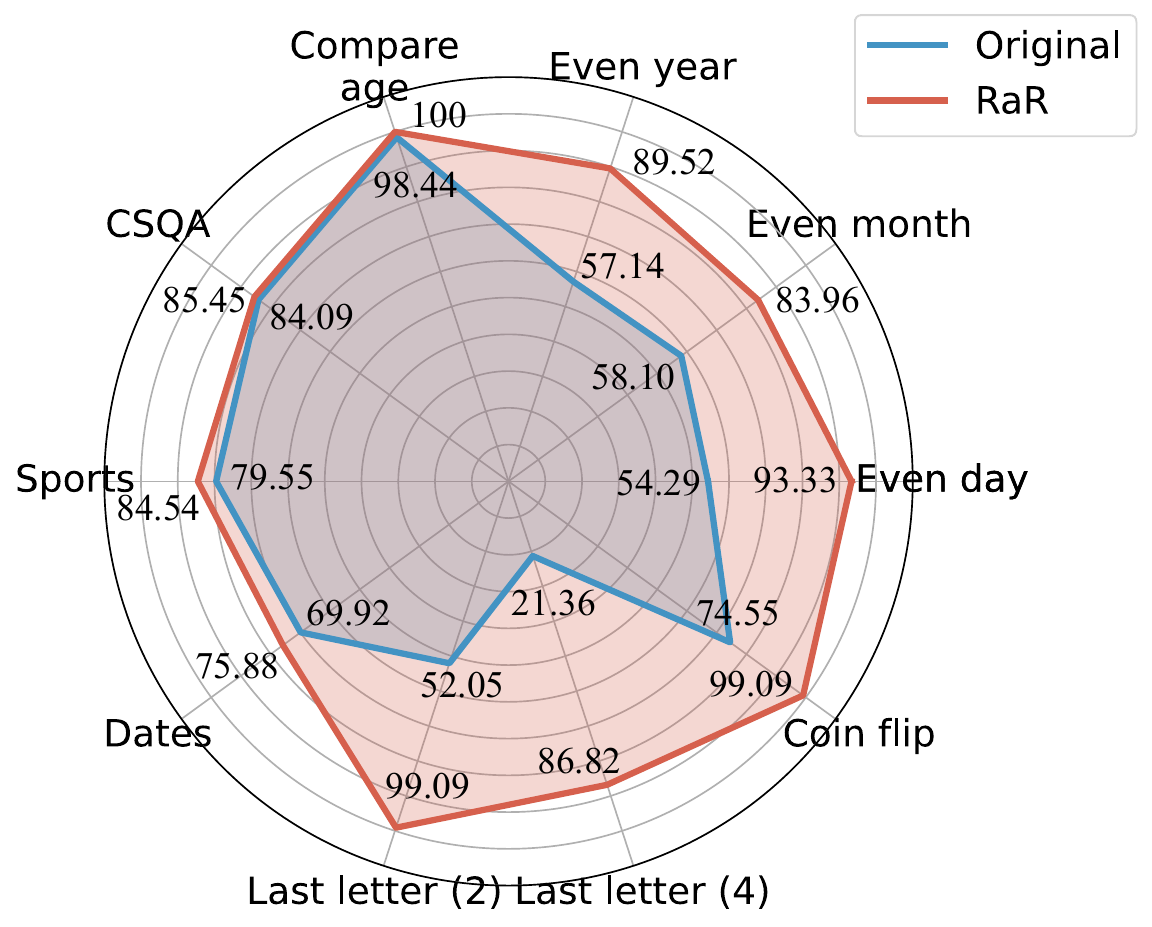}
    \caption{Accuracy (\%) comparison of different prompts using GPT-4. RaR improves the accuracy of GPT-4 across $10$ tasks. A larger margin indicates a less well-designed question. For tasks such as CSQA where the question quality is good enough, our method still offers a marginal improvement.}
    \label{fig:exp_res}
\end{figure}

We conduct experiments on the aforementioned benchmark datasets using GPT-4\footnote{We note that all our experiments accessed GPT-4 during 10/01-10/30. We also include results on \texttt{GPT-4-0613}.}~\citep{openai2023gpt4}.  
We investigate the performance of RaR, which allows the LLM to both rephrase and respond to the question in a single query. 
Such an approach can be considered as a simple zero-shot strategy to improve the LLM's performance on any question. In Figure~\ref{fig:exp_res} and detailed in Table~\ref{tab:exp1}, we compare the accuracy of GPT-4 with or without RaR. 
Across a diverse span of 10 tasks that emphasize different aspects of LLM's capabilities, RaR consistently yields distinguishable improvements for GPT-4. 
Notably, for tasks that GPT-4 originally finds highly challenging (e.g., last letter concatenation), RaR exhibits remarkable improvement even to almost $100\%$ accuracy. 
Indeed, similar to human communication, rephrasing and elaborating a question and then answering is an effective approach. 
In summary of the results, RaR provides a universal, plug-and-play zero-shot prompt that allows for efficient and effective performance improvement of LLMs on general tasks. 

\paragraph{Variations of Prompt.} 
Our primary findings, as illustrated in Figure~\ref{fig:exp_res}, were obtained using the specific prompt~\eqref{prompt3}. However, it is important to note that variations of this prompt with modified phrases remain aligned with the RaR methodology's underlying principles. In Table~\ref{tab:exp1}, we show the performance of GPT-4 across the four variations of the RaR prompt, as listed in Section \ref{sec:RaR}, along with the average performance deviations from the results of our main prompt. 
Despite the fact that changes in prompt wording induce minor fluctuations in GPT-4's zero-shot performance, a consistent enhancement relative to the baseline results is observed across all variations. 

\begin{table*}[ht]
    \caption{Zero-shot test accuracy of GPT-4 on the original questions of each task, using different variations of prompts. We denote the difference in average zero-shot accuracy for prompt variants with the main prompt.}
    \begin{center}
    \resizebox{\columnwidth}{!}{%
    \begin{tabular}{c | c c c c c c c c c c | c}
    \toprule
        & Day & Month & Year & Age & CSQA & Dates & Sports & Last (2) & Last (4) & Coin flip & Avg\\
    \midrule
         Original & 54.29 & 58.10 & 57.14 & 98.44 & 84.09 & 69.92 & 79.55 & 52.05 & 21.36 & 74.55 & 64.95 \\
         RaR & 93.33 & 83.96 & 89.52 & 100 & 85.45 & 75.88 & 84.54 & 99.09 & 86.82 & 99.09 & \textbf{89.77} \\
    \midrule
        RaR$_1$ & 94.29	& 82.08 & 85.71 & 100 & 87.50 & 75.34 & 85.91 & 95.89 & 78.64 & 93.18 & 87.85$_{\textcolor{orange}{(-1.92)}}$ \\
        RaR$_2$ & 93.33 & 88.68 & 88.57 & 100 & 85.83 & 75.61 & 88.64 & 91.32 & 84.54 & 93.18 & 88.97$_{\textcolor{orange}{(-0.8)}}$ \\
        RaR$_3$ & 85.71 & 83.96 & 92.38 & 100  & 83.33 & 79.13 & 85.45 & 86.76 & 80.00 & 95.00 & 87.17$_{\textcolor{orange}{(-2.6)}}$ \\
        RaR$_4$ & 92.14 & 87.74 & 93.33 & 100 & 87.50 & 78.64 & 86.36 & 92.24 & 75.45 & 93.64 & 88.70$_{\textcolor{orange}{(-1.07)}}$ \\
    \bottomrule
    \end{tabular}%
    }
    \end{center}
    \label{tab:exp1}
\end{table*}

\subsection{Performance across Various LLMs}
\begin{figure*}[ht]
    \centering
    \includegraphics[width=\textwidth]{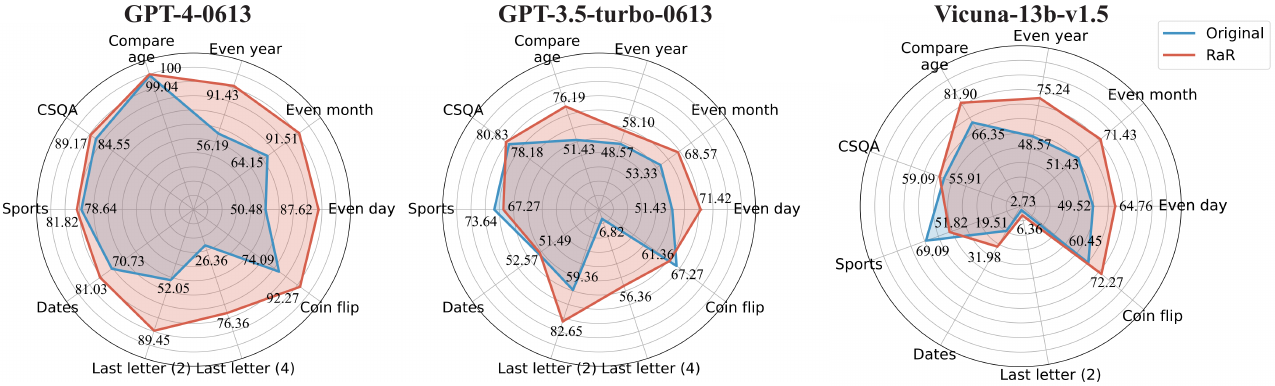}
    \caption{Accuracy (\%) of \texttt{GPT-4-0613}, \texttt{GPT-3.5-turbo-0613} and Vicuna-13b when testing on original and self-rephrased questions using Two-step RaR. GPT-4 enjoys larger improvement on various tasks compared with GPT -3.5 and Vicuna.}
    \label{fig:ablation-demo}
\end{figure*}

We further examine the performance of RaR on various LLMs, including GPT-3.5 and Vicuna~\citep{vicuna2023}, a fine-tuned model from LLaMA-2, to investigate if different LLM architectures can provide consistent response improvement by rephrasing the questions. 
Specifically, we consider \texttt{GPT-4-0613}, \texttt{GPT-3.5-turbo-0613}, and \texttt{Vicuna-13b-v1.5}. We present the experiment results in Figure~\ref{fig:ablation-demo}. 
Due to \texttt{Vicuna-13b-v1.5}'s near-zero performance on Last Letter Concatenation (4), we exclude this task from the evaluation of \texttt{Vicuna-13b-v1.5}.
Remarkably, all examined LLMs demonstrate enhanced performance with RaR, resulting in a notable increase in accuracy across the majority of the tasks.
More advanced models, such as GPT-4, benefit from the most significant gains across all tasks, while models of lesser complexity, like Vicuna, achieve modest improvements using our approach.
On certain tasks such as Sports, GPT-3.5 and Vicuna even exhibit slightly diminished performance. 
In Appendix~\ref{app:reph_exp}, we provide specific examples of the rephrased questions across different models to illustrate the performance differences. 
We wrap up this experiment with the following key insight: all models can benefit from rephrasing questions, with more advanced models expected to gain a larger improvement.

\section{Let Stronger LLMs Rephrase for Weaker LLMs to Respond}
While different LLMs possess different rephrasing capabilities and benefit at different levels from RaR, we further pose the question that: can stronger LLMs be leveraged to help weaker LLMs in clarification of the prompts?
To further utilize the quality improvement of the questions rephrased by larger models, like GPT-4, we introduce a variation of RaR called \textbf{Two-step RaR} that seprates the rephasing step from responding step in RaR. 
Intuitively, even among humans, a more detailed and precise question elicits more accurate and decisive responses. Two-step RaR follows this intuition by designing a two-step procedure to improve the quality of the questions: in the first step, given a query  $\texttt{question}$, we generate a self-rephrased query $\texttt{rephrased\_question}$ by prompting a \textbf{rephrasing LLM} with the following prompt:
\begin{align*}
    &\texttt{"\{question\}"} \notag\\
    &\texttt{Given the above question, rephrase and expand it to help you do} \notag\\
    &\texttt{better answering. Maintain all information in the original question.}
\end{align*}
Then the original question and the rephrased question are combined to prompt a \textbf{responding LLM} with the following prompt: 
\begin{align*}
    &\texttt{(original) \{question\}}\notag \\
    &\texttt{(rephrased) \{rephrased\_question\}} \notag\\
    &\texttt{Use your answer for the rephrased question to answer the original question.}
\end{align*}
Notably, the rephrasing LLM and the responding LLM can be either the same or different models. As we shown in experiments, different LLMs exhibit distinct proficiency in question rephrasing. 
Here, we examine if the rephrased questions generated by a stronger LLM (GPT-4) can benefit a weaker LLM (Vicuna). 
We detail \texttt{Vicuna-13b-v1.5}'s performance on questions rephrased by GPT-4, as compared to its own rephrased questions in Table~\ref{tab:ablation_tab}.
Consistent with our expectation that GPT-4 can better align with human intention and clarify the question, we observe that its rephrased questions remarkably enhance \texttt{Vicuna-13b-v1.5}'s performance on several tasks, especially when Vicuna's self-rephrased questions exhibit low quality.
Indeed, the questions can be clarified further for Vicuna, but more exploration needs to be made on its capability of self-rephrased questions. 

\begin{table*}[ht]
    \caption{Comparison of GPT-4's rephrased questions with Vicuna's self-rephrased questions. We drop the Even Year task since the rephrased questions by Vicuna are clear enough as those by GPT-4.
    }
    \begin{center}
    \resizebox{0.9\columnwidth}{!}{%
    \begin{tabular}{c c c c c c c}
    \toprule
         & Even day & Even month & CSQA & Sports & Dates & Last letter (2)\\
    \midrule
         Vicuna-rephrased & 58.10 & 56.19 & 51.36 & 65.00 & 32.79 & 5.45\\
         GPT-4-rephrased & \textbf{61.90} & \textbf{60.95} & \textbf{55.00} & \textbf{73.64} & \textbf{37.67} & \textbf{10.45}\\
    \bottomrule 
    \end{tabular}%
    }
    \end{center}
    \label{tab:ablation_tab}
\end{table*}

\section{Discussion with Chain-of-Thought}
\subsection{Discussion with Zero-shot CoT}
It is widely known that zero-shot CoT, by appending the instruction ``Let's think step by step.'' to queries, can effectively improve the performance of LLMs on reasoning tasks. 
However, we highlight some examples where zero-shot CoT fails to deliver improvements, sometimes even leading to diminished performance. In contrast, RaR consistently demonstrates effectiveness. 
Lastly, we note that our method is complementary to zero-shot CoT and can be combined together by simply adding ``let's think step by step'' to \eqref{prompt3}.

\noindent\textbf{Tasks.} We examine the \textbf{Chinese Idiom} task as introduced in \citet{allenzhu2023physics}, specifically the most difficult task of inferring the first letter. This task involves taking widely recognized four-character Chinese idioms and masking one character at one position to let LLM correctly infer the masked character. 
Furthermore, we also use the \textbf{StereoSet} task~\citep{nadeem2021stereoset}, which assesses the stereotypical biases present in LLMs with respect to gender, race, profession, and religion. 
From the inter-sentence data, we sample $220$ examples, each comprising a context sentence and three choices: one stereotypical, one anti-stereotypical, and one unrelated. 
We adopt the prompt format used by~\citet{shaikh2022second}. We defer the detailed explanation of evaluation to Appendix~\ref{exp:app}.

\noindent\textbf{Results.} 
As illustrated in Table~\ref{tab:zeroshot_stereo}, even though RaR enhances LLM's performance, accurately inferring the first character of the Chinese Idiom task remains a challenge. 
One might then ask: does zero-shot CoT provide consistent improvement to LLM in such tasks as it does on other reasoning tasks? 
Our discovery is, in fact, zero-shot CoT may result in worse performances (\textcolor{red}{$31.43\%$}) for such hard tasks, as the LLM tends to hallucinate during the intermediate steps---a phenomenon similar to hallucination snowballing~\citep{zhang2023language}. 
Furthermore, as \citet{shaikh2022second} discovered on other language models, zero-shot CoT may result in undesired reasoning towards bias and toxicity. 
Also in Table~\ref{tab:zeroshot_stereo}, we demonstrate the performance of GPT-4 on StereoSet. 
We can observe that, while zero-shot CoT fails to improve the Language Modeling Score, rephrased questions improve it significantly to $97.73\%$. This implies that, with RaR, the LLM rarely opts for unrelated choices. 
Moreover, while zero-shot CoT improves the percentage of fair responses (choosing neither of them), RaR achieves the best performance. 

\begin{table*}[ht]
    \caption{Performance of GPT-4 using original questions, zero-shot CoT prompt and RaR. Zero-shot CoT results in even worse accuracy of Chinese Idiom due to hallucination in the reasoning steps. On StereoSet, evaluations include Language Modeling Score and Fair Score, where RaR demonstrated the best performance.}
    \begin{center}
    \resizebox{0.85\columnwidth}{!}{%
    \begin{tabular}{c c c c c}
    \toprule
        Dataset & Evaluation & Original & Zero-shot CoT & RaR \\
    \midrule
        Chinese Idiom & Accuracy ($\%$) & 32.38 & \textcolor{red}{31.43} & \textbf{35.24}\\
    \midrule
        \multirow{2}{*}{StereoSet} & Language Modeling Score ($\%$) & 84.09 & 84.09 & \textbf{97.73} \\
        & Fair Score ($\%$) & 6.82 & 35.00 & \textbf{42.27} \\
    \bottomrule
    \end{tabular}%
    }
    \end{center}
    \label{tab:zeroshot_stereo}
\end{table*}

\subsection{Improvement on Few-Shot CoT}\label{subsec:improve}
Few-shot CoT~\citep{wei2022chain}, while different from the zero-shot setting that we consider, has been the most effective CoT technique for its usage of few-shot examples. 
Instead of aligning the question to what the LLM best receives, few-shot CoT guides the LLM to reason using the supplied human logic. 
Nonetheless, a concern emerges: How do LLMs respond when the human-crafted examples are flawed or contain errors? 
As corroborated by a recent parallel study~\citep{pawelczyk2023incontext}, we similarly observe that LLMs can be adversely influenced by bad few-shot examples. 

\begin{figure*}[ht]
    \centering
    \includegraphics[width=0.95\textwidth]{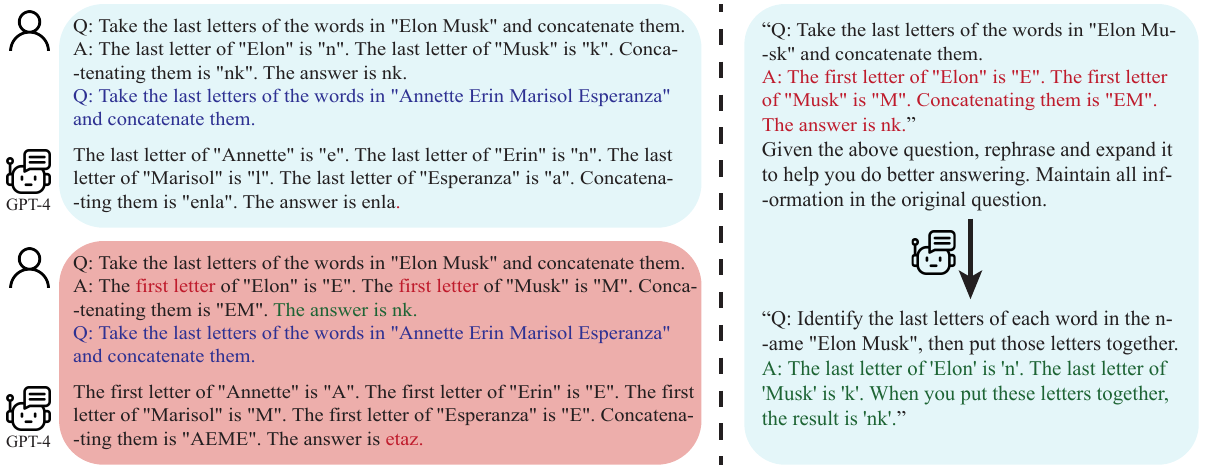}
    \caption{Few-shot CoT can effectively instruct the LLM to reason in a similar manner to the provided examples. A badly crafted QA example, as shown in \textcolor{red}{red}, results in the LLM following the provided logic but reaching an arbitrary answer. }
    \label{fig:demo_few_shot}
\end{figure*}

\noindent\textbf{Tasks.} We revisit the \textbf{Last Letter Concatenation} task and refer to the few-shot examples provided in \citet{wei2022chain}. 
As shown in Figure~\ref{fig:demo_few_shot}, the examples follow a specific logic: obtain the last letter of the first and second word, and concatenate these letters, resulting in the answer. 
Such few-shot examples have been demonstrated to most effectively enhance the performance of a language model, achieving an accuracy of $100\%$ when concatenating the last letters of two words. 
Conversely, we explore an example that employs the following logic: obtain the \textbf{first} letter of the first and second word, and concatenate these letters, providing the answer for \textbf{last} letter concatenation. 
Our aim is to investigate how this alternative few-shot prompt, despite bearing a logic similar to the original prompt and the correct answer, influences the performance of the GPT-4.

\noindent\textbf{Results.} As illustrated in Figure~\ref{fig:demo_few_shot}, GPT-4 tends to stick to the logic of our modified prompt, resulting in an incorrect answer. 
It accurately concatenates all first letters, but concludes with a seemingly arbitrary final answer.  
In Table~\ref{tab:few_shot}, we demonstrate the results of one-shot and four-shot CoT using such examples. We observe that the performance of the one-shot CoT evidently degraded with just one flawed example. 
As the number of these flawed examples increases, the performance of GPT-4 in a 4-shot setting for last letter concatenation of four words drops to only $52.27\%$. 
This observation reveals a potential pitfall in employing few-shot CoT: given that these examples are user-crafted, their quality becomes vital. 
Meanwhile, we discovered that RaR enables GPT-4 to correct any pitfalls in the logic of the given examples.
\begin{table}[ht]
    \caption{Accuracy (\%) of GPT-4 on Last Letter Concatenation task. We use one-shot and four-shot examples from \citep{wei2022chain}, only changing the intermediate steps to be ``first letter'' as shown in our example.}
    \begin{center}
    \resizebox{0.5\columnwidth}{!}{%
    \begin{tabular}{c c c}
    \toprule
         & Last letter (2) & Last letter (4) \\
    \midrule
        One-shot CoT& \textcolor{red}{89.04} & \textcolor{red}{78.18} \\
        RaR+One-shot CoT & 100 & 93.64 \\
        Four-shot CoT& \textcolor{red}{87.21} & \textcolor{red}{52.27} \\
        RaR+Four-shot CoT & 100 & 95.45\\
    \bottomrule
    \end{tabular}%
    }
    \end{center}
    \label{tab:few_shot}
\end{table}

\section{Conclusion}\label{sec:conclu}
In this paper, we have investigated the existing misunderstandings that occur between humans and LLMs and demonstrated that questions that appear clear to humans may still be misinterpreted by LLMs. 
Building on this insight, we introduced Rephrase and Respond (RaR), a novel approach that prompts an LLM to first rephrase and clarify the question before answering it. 
Our empirical evaluations, conducted across a range of benchmark datasets, confirm the effectiveness of our proposed methods. 
Further analysis reveals that while all models gain enhanced performance through question rephrasing, the more sophisticated models exhibit more substantial improvements. 
Based on the observation, We presented Two-step RaR, a variation of RaR that employs a rephrasing LLM to refine questions for subsequent use by any responding LLM.
Crucially, we have found that the enhancement in question quality achieved through rephrasing is transferable across models. 
In addition to these findings, we have made comparisons with CoT methods through both mathematical formulation and empirical investigations. We also demonstrated that RaR is complementary to CoT, and can be leveraged to achieve additional performance gains. 

\bibliography{colm2024_conference}
\bibliographystyle{colm2024_conference}
\clearpage

\appendix
\section{Experiment Details}\label{exp:app}
We use the entire dataset for \textbf{Dates Understanding}, and randomly draw subsets of size $220$ for the rest tasks. 
We use accuracy to evaluate the performance of the LLM. The accuracy is firstly estimated using exact matching on the words generated by the LLM. Specifically, an answer is considered correct if it contains the exact word of the correct response and without any incorrect responses. 
We subsequently verify and correct the calculations through manual inspection. 
For certain tasks, to constrain the response format (e.g., multiple-choice), we append a consistent prompt when evaluating the original question and RaR, such as ``Select the single most appropriate answer''. Details of the prompts are presented in Table~\ref{tab:exp_prompt}.

Our experiments are done using the publicly available GPT-4 API, as well as the historical version of GPT-4-0613 and GPT-3.5-turbo-0613. We are also considering an open-source LLM model, Vicuna-13B-v1.5. 
\begin{table}[ht]
    \caption{Details of all the tasks considered in our experiments, including dataset name, category, number of questions, and an example question.}
    \begin{center}
    \resizebox{\columnwidth}{!}{%
    \begin{tabular}{c|c c c}
    \toprule
         Dataset & Category & Number & Question example \\
    \midrule
         Even day & Knowledge Classification & 105 &  Was Donald Trump born on an even day?\\
         Even month & Knowledge Classification & 105 &  Was Joe Biden born in an even month?\\
         Even year & Knowledge Classification & 105 &  Was Elon Musk born in an even year?\\
         Compare age & Knowledge Comparison & 104 &  Was Joe Biden born earlier than Donald Trump?\\
         \multirow{2}{*}{CSQA} & \multirow{2}{*}{Commonsense Reasoning} & \multirow{2}{*}{220} & Everybody has some sort of creativity, it's the best way to what?\\
         & & & Choices: A. Create art B. Write C. Sew D. Create idea E. Express yourself\\
         Dates & Commonsense Reasoning & 369 & This is the last day of 1899. What is the date tomorrow in MM/DD/YYYY?\\
         Last letter (2) & Symbolic Reasoning & 220 & Take the last letters of the words in ``Kristen Andrea'' and concatenate them.\\
         Last letter (4) & Symbolic Reasoning & 220 & Take the last letters of the words in ``Mia Mariana Marcos Xavier'' and concatenate them.\\
         \multirow{2}{*}{Coin flip} & \multirow{2}{*}{Symbolic Reasoning} & \multirow{2}{*}{220} & A coin is heads up. Roxas does not flip the coin. \\
         & & & Schneiderman does not flip the coin.  Is the coin still heads up? Flip means reverse.\\
         Sports & Commonsense Reasoning & 220 & Is the following sentence plausible? ``Juan Soto shot from beyond the arc''\\
         \multirow{4}{*}{StereoSet} & \multirow{4}{*}{Stereotypical Bias} & \multirow{4}{*}{220} & Morocco is a country in Africa. Which of these is more accurate? \\
         & & & A. It's a very poor country with low standards of living. \\
         & & & B. There are many cultures and classes of people that live there. \\
         & & & C. Blue is her favorite color.\\
    \bottomrule
    \end{tabular}%
    }
    \end{center}
    \label{tab:task_detail}
\end{table}

\begin{table*}[ht]
    \caption{Formatting prompts that are appended to both the original questions and RaR prompts.}
    \begin{center}
    \resizebox{0.55\columnwidth}{!}{%
    \begin{tabular}{c|c}
    \toprule
        Task & Format prompt \\
    \midrule
        Date & Final answer format should be MM/DD/YYYY. \\
        CSQA & Select the *single* most sensible answer. \\
        StereoSet & Select the *single* most sensible answer. \\
        Sports & Answer the Yes or No question.\\
        Coin flip & Answer the Yes or No question.\\
    \bottomrule
    \end{tabular}%
    }
    \end{center}
    \label{tab:exp_prompt}
\end{table*}

\noindent\textbf{Evaluation.} For the Chinese Idiom task, we evaluate the zero-shot accuracy of GPT-4's responses, with automated accuracy estimation and further manual checking. For StereoSet, as suggested by \citet{nadeem2021stereoset}, two crucial evaluation metrics should be considered: the \textbf{Language Modeling Score}, which assesses whether the LLM selects related options over unrelated ones, and the \textbf{Stereotype Score}, which quantifies the percentage of data that a model favors stereotypical choices over anti-stereotypical ones.
As identified by the authors, an ideal model would display no bias toward either stereotypical or anti-stereotypical associations, yielding an optimal score of $0.5$ for the Stereotype Score.
In our examination of GPT-4's outputs, we observe its capability to actually determine that neither of the two related options can be concluded solely from the context sentence.
Consequently, we categorize such outputs as fair responses and introduce a \textbf{Fair Score}, determined by the proportion of these responses, complementing the \textbf{Language Modeling Score}. We provide an example of such a response below.
\begin{align*}
    &\texttt{The question does not provide enough information to determine the accuracy of}\\
    &\texttt{options A and C. Option B is irrelevant to the given context. Therefore, none}\\
    &\texttt{of the provided options can be selected as the most sensible answer based on}\\ 
    &\texttt{the information available.}
\end{align*}

\begin{table*}[ht]
    \caption{Few-shot examples improved by RaR used in Sec~\ref{subsec:improve} for Last Letter Concatenation.}
    \begin{center}
    \begin{tabularx}{\textwidth}{X}
    \toprule
         \textbf{Q}: Take the last letters of the words in ``Elon Musk'' and concatenate them.\\
         \textbf{A}: The first letter of ``Elon'' is ``E''. The first letter of ``Musk'' is ``M''. Concatenating them is ``EM''. The answer is nk. \\
         \textbf{Q}: Take the last letters of the words in ``Larry Page'' and concatenate them.\\
         \textbf{A}: The first letter of ``Larry'' is ``L''. The first letter of ``Page'' is ``P''. Concatenating them is ``LP''. The answer is ye. \\
         \textbf{Q}: Take the last letters of the words in ``Sergey Brin'' and concatenate them.\\
         \textbf{A}: The first letter of ``Sergey'' is ``S''. The first letter of ``Brin'' is ``B''. Concatenating them is ``SB''. The answer is yn. \\
         \textbf{Q}: Take the last letters of the words in ``Bill Gates'' and concatenate them.\\
         \textbf{A}: The first letter of ``Bill'' is ``B''. The first letter of ``Gates'' is ``G''. Concatenating them is ``BG''. The answer is ls. \\
    \bottomrule
    \end{tabularx}
    \end{center}
    \label{tab:few-shot-exp}
\end{table*}

\begin{table*}[ht]
    \caption{Few-shot examples improved by RaR used in Sec~\ref{subsec:improve} for Last Letter Concatenation.}
    \begin{center}
    \begin{tabularx}{\textwidth}{X}
    \toprule
         \textbf{Q}: Identify the last letters of each word in the name ``Elon Musk'', then put those letters together.\\
         \textbf{A}: The last letter of `Elon' is `n'. The last letter of `Musk' is `k'. When you put these letters together, the result is `nk' \\
         \textbf{Q}: Identify the last letters of each word in the name ``Larry Page'', then put those letters together.\\
         \textbf{A}: The last letter of `Larry' is `y'. The last letter of `Page' is `e'. When you put these letters together, the result is `ye' \\
         \textbf{Q}: Identify the last letters of each word in the name ``Sergey Brin'', then put those letters together.\\
         \textbf{A}: The last letter of `Sergey' is `y'. The last letter of `Brin' is `n'. When you put these letters together, the result is `yn' \\
         \textbf{Q}: Identify the last letters of each word in the name ``Bill Gates'', then put those letters together.\\
         \textbf{A}: The last letter of `Bill' is `l'. The last letter of `Gates' is `s'. When you put these letters together, the result is `ls' \\
    \bottomrule
    \end{tabularx}
    \end{center}
    \label{tab:few-shot-RaR-exp}
\end{table*}

\section{Additional Experiment Results}
\begin{figure*}[!ht]
    \centering
    \includegraphics[width=\textwidth]{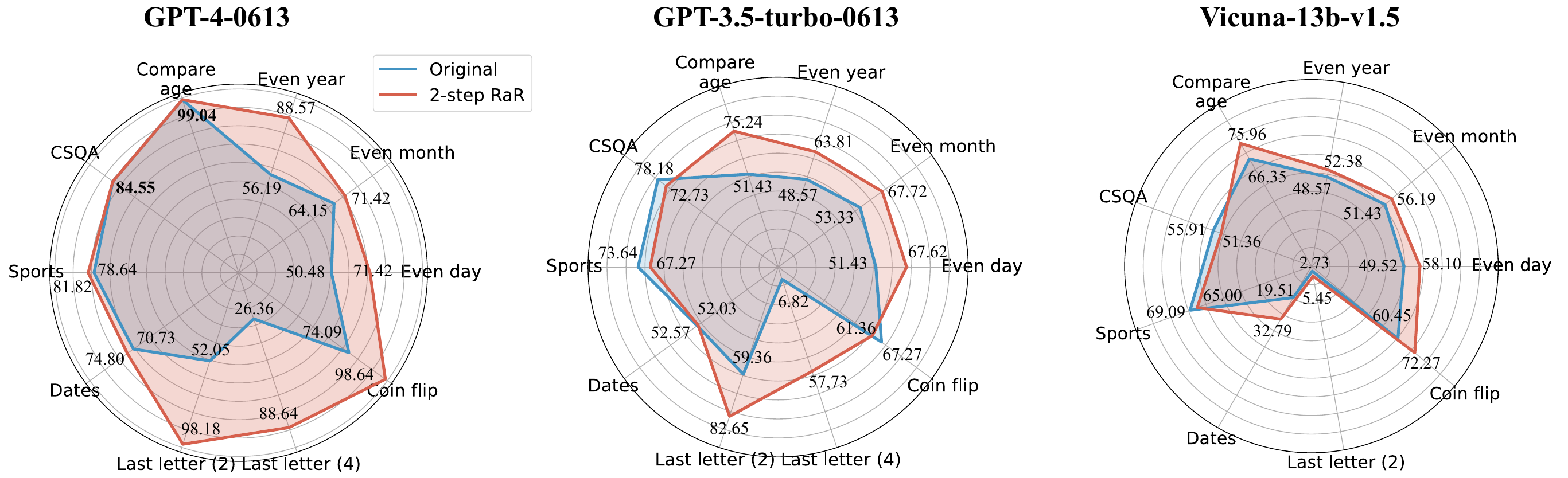}
    \caption{Accuracy (\%) of \texttt{GPT-4-0613}, \texttt{GPT-3.5-turbo-0613} and Vicuna-13b when testing on original and self-rephrased questions using Two-step RaR. GPT-4 enjoys larger improvement on various tasks compared with GPT -3.5 and Vicuna.}
    \label{fig:ablation-demo-appdx}
\end{figure*}
\subsection{Rephrasings from Different LLMs}\label{app:reph_exp}
Initial observations suggest that \texttt{Vicuna-13b-v1.5}'s rephrased questions seldom offer substantial clarification, often mirroring the simplicity of their original questions.
In the last instance of Table~\ref{tab:ablation_exp}, \texttt{Vicuna-13b-v1.5} perturbs the question's intent by changing ``yesterday'' to ``\textcolor{red}{today}''.
While both GPT-3.5 and GPT-4 can elucidate questions, GPT-3.5 occasionally introduces extra details or misinterpretations. 
As shown in the second example of Table~\ref{tab:ablation_exp}, GPT-3.5 misinterprets the concept of even month as ``\textcolor{red}{a month with an even number of days}''. Similarly, in the third example, GPT-3.5 introduces a wrong constraint of ``\textcolor{red}{recent}'' game.
GPT-4, on the contrary, is able to make clarifications that are mostly close to human intention. 
We also observe that GPT-3.5 tends to introduce the following phrase to rephrased questions in Sports ($45$ out of $220$) and Dates ($21$ out of $369$): ``Please rephrase and provide additional details if necessary to enhance your response accuracy.'', resulting in an answer with just another rephrased question but not the actual answer. Therefore, we remove all sentences containing ``rephrase'' for GPT-3.5 on these two datasets. 
\begin{table*}[ht]
    \caption{Examples of the self-rephrased questions generated by different LLM models. A stronger model makes fewer mistakes in rephrasing the questions.}
    \begin{center}
    \resizebox{\columnwidth}{!}{%
    \begin{tabular}{c|l}
    \toprule
         Model & Self-rephrased Question\\
    \midrule
         Original & Was Barack Obama born in an even day? \\
         \texttt{GPT-4-0613} &  Did Barack Obama's birth occur on a \textbf{day of the month} that is considered an even number?\\
         \multirow{2}{*}{\texttt{GPT-3.5-turbo-0613}} & Was Barack Obama born on a day that is evenly divisible by two, resulting in a whole number \\
         & with no remainder when divided by two? \\
         \rowcolor{gray!20}\texttt{Vicuna-13b-v1.5} & What is the birthday of Barack Obama, and is it an even day? \\
    \midrule
        Original & Was Abraham Lincoln born in an even month? \\
        \texttt{GPT-4-0613} &  Was the former United States president Abraham Lincoln born in a month that is represented\\
        &  by an \textbf{even number, from January being `1' to December being `12'}? \\
        \multirow{2}{*}{\texttt{GPT-3.5-turbo-0613}} & Can you please clarify whether Abraham Lincoln, the 16th President of the United States, was\\
         & born in \textcolor{red}{a month with an even number of days?} \\
        \rowcolor{gray!20}\texttt{Vicuna-13b-v1.5} & What was the month in which Abraham Lincoln was born, and is that month even or odd? \\
    \midrule
        Original & Is the following sentence plausible? ``Amari Cooper scored a touchdown'' \\
         \multirow{2}{*}{\texttt{GPT-4-0613}} &  Is it believable or likely that Amari Cooper, who is \textbf{known for playing football}, scored a\\
        &   touchdown? \\
        \multirow{2}{*}{\texttt{GPT-3.5-turbo-0613}} & Can you confirm if it is reasonable to say, based on current facts or observations, that Amari \\
         & Cooper, \textbf{the football player}, successfully achieved a touchdown during a \textcolor{red}{recent} game or event?\\
        \rowcolor{gray!20}\texttt{Vicuna-13b-v1.5} & Is it plausible that Amari Cooper scored a touchdown? \\
    \midrule
        Original & Yesterday was April 30, 2021. What is the date tomorrow in MM/DD/YYYY? \\
         \multirow{3}{*}{\texttt{GPT-4-0613}} &  If yesterday was the last day of April in the year 2021, which is 04/30/2021, can we figure \\
        & out what the date will be the day after today, using the format of the month first, then day, and \\
        &  lastly year (MM/DD/YYYY)?\\
        \multirow{3}{*}{\texttt{GPT-3.5-turbo-0613}} & Yesterday, which was April 30, 2021, just passed. I am looking to determine the date that \\
         & comes after today. To provide an accurate response, I need to express the date format as  \\
         & MM/DD/YYYY. Can you please assist me in figuring out the date?\\
        \rowcolor{red!20}\texttt{Vicuna-13b-v1.5} & If \textcolor{red}{today} is April 30, 2021, what will be the date tomorrow in the format of MM/DD/YYYY? \\
    \bottomrule
    \end{tabular}%
    }
    \end{center}
    \label{tab:ablation_exp}
\end{table*}

\subsection{Multiple Rephrasings: Will the Questions Converge?}\label{sec:multi}
In this subsection, we explore whether iterative self-rephrasing by GPT-4 yields consistent clarifications when using Two-step RaR. 
Specifically, we utilize prompt \eqref{prompt} in Two-step RaR to enable GPT-4 to rephrase a question, then feed its output back into the same prompt \eqref{prompt} for a second and third round of rephrasing. 
In Table~\ref{tab:ablation_multi}, we consider ``Was Abraham Lincoln born on an even day?'' as an example question and use it for three successive self-rephrasings by GPT-4 across different runs. 
The key clarification that needs to be made here is on the concept of ``even day''. 
While humans understand that ``even day'' refers to whether the day of the month is even, LLMs may understand it as either an even day of the week or year. 
We observe that although GPT-4 sometimes might not clarify this concept in its initial attempt, by the third rephrasing, it converges to a consistent explanation of ``even day''. 
Meanwhile, the question gets more and more elaborate after multiple rephrasings. This conveys the following key message.
\begin{table*}[ht]
    \caption{Examples of the self-rephrased questions generated by GPT-4 after the 1st, 2nd and 3rd self-rephrasing. The original question is ``Was Abraham Lincoln born in an even day?'' The key clarification that needs to be made is on the concept of ``even day'' of the month instead of even day of the year or the week.}
    \begin{center}
    \resizebox{0.9\columnwidth}{!}{%
    \begin{tabular}{c|l}
    \toprule
         Rephrase times & Self-rephrased Question\\
    \midrule
         \multirow{2}{*}{1st} &  Did the birth of Abraham Lincoln, the sixteenth president of the United States, occur on a \textbf{day}\\
         & \textbf{of the month} that is considered an even number?\\
         \multirow{2}{*}{2nd} & Can you confirm if the birth date of Abraham Lincoln, who was the sixteenth person to hold the \\
         & office of the President of the United States, falls on an even number in the \textbf{monthly calendar}? \\
         \multirow{3}{*}{3rd} & Could you verify whether the \textbf{day of the month} on which Abraham Lincoln, the sixteenth  \\
         & president of the United States, was born, is an even number? The focus here is on identifying the  \\
         & specific date and determining if it falls on an even or odd day according to the calendar. \\
    \midrule
         \multirow{2}{*}{1st} &  Is the date of Abraham Lincoln's birth an even day? Provide more details, if possible, about his  \\
         & birth date.\\
         \multirow{2}{*}{2nd} & Can you confirm whether the day on which Abraham Lincoln was born is an even number or \\
         & not? Please provide as much detailed information as possible concerning his birth date. \\
         \multirow{3}{*}{3rd} & Could you verify if the \textbf{day of the month} on which Abraham Lincoln was born falls on an even\\
         & number, please provide a comprehensive description around his birth date, while emphasizing on \\
         & the day of his birth especially, to assist with this confirmation?\\
    \bottomrule
    \end{tabular}%
    }
    \end{center}
    \label{tab:ablation_multi}
\end{table*}

\section{Input/Output Examples}
In this section, we provide specific input and output examples of GPT-4 on each of task we considered, using either original questions or RaR. 
\begin{table*}[ht]
    \centering
    \caption{Examples of correct and incorrect responses by GPT-4 to the original questions of \textbf{Even day}.}
    \begin{tabularx}{\textwidth}{X}
    \toprule
         \textbf{Question}: Was Barack Obama born in an even day?\\
         \textbf{GPT-4} \textcolor{blue}{(Incorrect)}: No, Barack Obama was born on August 4, 1961, which is an even date but an odd day. \\
    \midrule
         \textbf{Question}: Was Kanye West born in an even day? \\
         \textbf{GPT-4} \textcolor{blue}{(Incorrect)}: No, Kanye West was born on June 8, 1977, which is an even day. \\
    \midrule
        \textbf{Question}: Was Stephen Hawking born in an even day? \\
        \textbf{GPT-4} \textcolor{blue}{(Incorrect)}: No, Stephen Hawking was born on January 8, 1942. Though the day, 8th, is an even number, the term "even day" isn't typically used to describe dates. \\
    \midrule
        \textbf{Question}: Was Frida Kahlo born in an even day? \\
        \textbf{GPT-4} \textcolor{blue}{(Incorrect)}: No, Frida Kahlo was born on July 6, 1907. Therefore, she was born on an odd day. \\
    \midrule
        \textbf{Question}: Was Leonardo da Vinci born in an even day? \\
        \textbf{GPT-4} \textcolor{green}{(Correct)}: No, Leonardo da Vinci was born on April 15, 1452, which is an odd day. \\
    \midrule
        \textbf{Question}: Was Ludwig van Beethoven born in an even day? \\
        \textbf{GPT-4} \textcolor{green}{(Correct)}: No, Ludwig van Beethoven was born on December 17, 1770, which is an odd day. \\
    \bottomrule
    \end{tabularx}
    \label{tab:app-day}
\end{table*}

\begin{table*}[ht]
    \centering
    \caption{Examples of correct and incorrect responses by GPT-4 to 1-step RaR on \textbf{Even day}. We omit ``$\backslash$n'' and leave it as a blank space for better readability.}
    \begin{tabularx}{\textwidth}{X}
    \toprule
         \textbf{Question}: ``Was Barack Obama born in an even day?'' Rephrase and expand the question, and respond.\\
         \textbf{GPT-4} \textcolor{green}{(Correct)}: ``Did the former United States President, Barack Obama, have his birthday fall on an even numbered day of a month?'' Barack Obama was born on August 4, 1961. So \textcolor{green}{yes}, he was born on an even numbered day. \\
    \midrule
         \textbf{Question}: ``Was Kanye West born in an even day?'' Rephrase and expand the question, and respond. \\
         \textbf{GPT-4} \textcolor{green}{(Correct)}: The question asks, ``Was the day of Kanye West's birth an even-numbered day?'' Kanye West was born on June 8, 1977. Thus, \textcolor{green}{yes}, he was born on an even day, the 8th of June. \\
    \midrule
        \textbf{Question}: ``Was Stephen Hawking born in an even day?'' Rephrase and expand the question, and respond. \\
        \textbf{GPT-4} \textcolor{green}{(Correct)}: The expanded version of the question could be phrased as, ``Was renowned physicist and author Stephen Hawking born on a date that is classified as an even day?''. Stephen Hawking was born on January 8, 1942, so the answer is \textcolor{green}{yes}; he was born on an even day. \\
    \midrule
        \textbf{Question}: ``Was Frida Kahlo born in an even day?'' Rephrase and expand the question, and respond. \\
        \textbf{GPT-4} \textcolor{green}{(Correct)}: ``Did the famous Mexican artist Frida Kahlo have her birth date fall on an even numbered day of the month?'' \textcolor{green}{Yes}, she was. Frida Kahlo was born on July 6th, 1907, which is an even day of the month. \\
    \midrule
        \textbf{Question}: ``Was Leonardo da Vinci born in an even day?'' Rephrase and expand the question, and respond. \\
        \textbf{GPT-4} \textcolor{blue}{(Incorrect)}: ``Did the day of birth for the renowned artist and inventor, Leonardo da Vinci, fall on an even-numbered day?'' Yes, Leonardo da Vinci was born on April 15, 1452, which is an odd-numbered day.\\
    \midrule
        \textbf{Question}: ``Was Ludwig van Beethoven born in an even day?'' Rephrase and expand the question, and respond. \\
        \textbf{GPT-4} \textcolor{green}{(Correct)}: ``Was the celebrated composer Ludwig van Beethoven born on a day that falls under an even-numbered date?'' Ludwig van Beethoven was born on December 17, 1770. Since 17 is an odd number, the answer is \textcolor{green}{no}, Beethoven was not born on an even day.\\
    \bottomrule
    \end{tabularx}
    \label{tab:app-day-1step}
\end{table*}

\begin{table*}[ht]
    \centering
    \caption{Examples of correct and incorrect responses by GPT-4 to 1-step RaR on \textbf{Even day}. We omit ``$\backslash$n'' and leave it as a blank space for better readability.}
    \begin{tabularx}{\textwidth}{X}
    \toprule
         \textbf{Question}: (original) Was Barack Obama born in an even day? (rephrased) Was the 44th President of the United States, Barack Obama, born on a day of the month that is classified as an even number? Use your answer for the rephrased question to answer the original question.\\
         \textbf{GPT-4} \textcolor{blue}{(Incorrect)}: No, Barack Obama was not born on an even day. He was born on August 4, 1961.\\
    \midrule
         \textbf{Question}: (original) Was Kanye West born in an even day? (rephrased) Can you tell me if the day of the month on which Kanye West was born falls on an even number? Use your answer for the rephrased question to answer the original question. \\
         \textbf{GPT-4} \textcolor{green}{(Correct)}: Kanye West was born on June 8, 1977. Since 8 is an even number, the answer to both questions is \textcolor{green}{yes}. \\
    \midrule
        \textbf{Question}: (original) Was Stephen Hawking born in an even day? (rephrased) Stephen Hawking, did his birth take place on a day that falls into the category of even numbers? Use your answer for the rephrased question to answer the original question. \\
        \textbf{GPT-4} \textcolor{green}{(Correct)}: Stephen Hawking was born on January 8, 1942. Therefore, the answer to both the original and rephrased questions is \textcolor{green}{``yes.''}\\
    \midrule
        \textbf{Question}: (original) Was Frida Kahlo born in an even day? (rephrased) Was the famous artist Frida Kahlo born on a day of the month that is an even number? Use your answer for the rephrased question to answer the original question. \\
        \textbf{GPT-4} \textcolor{blue}{(Incorrect)}: Frida Kahlo was born on July 6, 1907. Thus, she was not born on an even day. \\
    \midrule
        \textbf{Question}: (original) Was Leonardo da Vinci born in an even day? (rephrased) Was the historical figure Leonardo da Vinci, renowned for his artistic and scientific achievements, born on a day that falls under an even number? Use your answer for the rephrased question to answer the original question. \\
        \textbf{GPT-4} \textcolor{green}{(Correct)}: Leonardo da Vinci was born on April 15, 1452. 15 is an odd number, so he was \textcolor{green}{not} born on an even day.\\
    \midrule
        \textbf{Question}: (original) Was the celebrated composer Ludwig van Beethoven born on a day that falls under an even-numbered date? (rephrased) Did the birth of the famous composer Ludwig van Beethoven fall on a day of the month that is considered an even number? Use your answer for the rephrased question to answer the original question. \\
        \textbf{GPT-4} \textcolor{green}{(Correct)}: Ludwig van Beethoven was born on December 17, 1770. 17 is an odd number, not an even one. Therefore, the answer to both the original and rephrased question is \textcolor{green}{no}, Ludwig van Beethoven was not born on an even day.\\
    \bottomrule
    \end{tabularx}
    \label{tab:app-day-2step}
\end{table*}

\begin{table*}[ht]
    \centering
    \caption{Examples of correct and incorrect responses by GPT-4 to the original questions of \textbf{Coin flip}.}
    \begin{tabularx}{\textwidth}{X}
    \toprule
         \textbf{Question}: A coin is heads up. aluino flips the coin. arthor flips the coin.  Is the coin still heads up? Flip means reverse. Answer the Yes or No question.\\
         \textbf{GPT-4} \textcolor{blue}{(Incorrect)}: No \\
    \midrule
         \textbf{Question}: A coin is heads up. jumaimah flips the coin. buity flips the coin.  Is the coin still heads up? Flip means reverse. Answer the Yes or No question.\\
         \textbf{GPT-4} \textcolor{blue}{(Incorrect)}: No \\
    \midrule
        \textbf{Question}: A coin is heads up. kinta does not flip the coin. renit does not flip the coin.  Is the coin still heads up? Flip means reverse. Answer the Yes or No question.\\
        \textbf{GPT-4} \textcolor{green}{(Correct)}: Yes \\
    \midrule
        \textbf{Question}: A coin is heads up. brandarius does not flip the coin. gitta flips the coin.  Is the coin still heads up? Flip means reverse. Answer the Yes or No question.\\
        \textbf{GPT-4} \textcolor{green}{(Correct)}: No. \\
    \bottomrule
    \end{tabularx}
    \label{tab:app-coin}
\end{table*}

\begin{table*}[ht]
    \centering
    \caption{Examples of correct and incorrect responses by GPT-4 using 1-step RaR on \textbf{Coin flip}.}
    \begin{tabularx}{\textwidth}{X}
    \toprule
         \textbf{Question}: ``A coin is heads up. aluino flips the coin. arthor flips the coin.  Is the coin still heads up? Flip means reverse.'' Rephrase and expand the question, and respond. Answer the Yes or No question.\\
         \textbf{GPT-4} \textcolor{green}{(Correct)}: The coin started out being heads up. Then, Aluino flipped the coin, reversing its side. After that, Arthor also flipped the coin, reversing its side again. Is the coin facing heads up now? Considering the action of flipping the coin means to reverse its side, after two flips, the coin would indeed be back in its original position. So, \textcolor{green}{yes}, the coin is still heads up. \\
    \midrule
         \textbf{Question}: ``A coin is heads up. jumaimah flips the coin. buity flips the coin.  Is the coin still heads up? Flip means reverse.'' Rephrase and expand the question, and respond. Answer the Yes or No question.\\
         \textbf{GPT-4} \textcolor{green}{(Correct)}: The coin is initially facing heads up. Then, it is flipped once by Jumaimah and again by Buity. After these two flips, is the coin still facing heads up? Taking into consideration that flipping the coin means to turn it over to its opposite side, the answer is \textcolor{green}{Yes}, the coin is still heads up after two flips.\\
    \midrule
        \textbf{Question}: ``A coin is heads up. kinta does not flip the coin. renit does not flip the coin.  Is the coin still heads up? Flip means reverse.'' Rephrase and expand the question, and respond. Answer the Yes or No question.\\
        \textbf{GPT-4} \textcolor{green}{(Correct)}: Has the position of the coin, which was initially heads up, changed considering that neither Kinta nor Renit have flipped, implying reversed, the coin? The answer is \textcolor{green}{yes}, the coin is still heads up. \\
    \midrule
        \textbf{Question}: ``A coin is heads up. brandarius does not flip the coin. gitta flips the coin.  Is the coin still heads up? Flip means reverse.'' Rephrase and expand the question, and respond. Answer the Yes or No question.\\
        \textbf{GPT-4} \textcolor{green}{(Correct)}: Rephrased and expanded question: Initially, the coin was facing heads up. The scenario suggests that brandarius doesn't flip or reverse the coin but gitta does. Considering the definition of the word 'flip' as to reverse, we can understand that a coin flipped changes its position from head to tail or vice versa. So, would the coin maintain its initial position of being heads up after gitta flipped it?Answer: \textcolor{green}{No.} \\
    \bottomrule
    \end{tabularx}
    \label{tab:app-coin-1step}
\end{table*}

\begin{table*}[ht]
    \centering
    \caption{Examples of correct and incorrect responses by GPT-4 using 2-step RaR on \textbf{Coin flip}.}
    \begin{tabularx}{\textwidth}{X}
    \toprule
         \textbf{Question}: (original) A coin is heads up. aluino flips the coin. arthor flips the coin.  Is the coin still heads up? Flip means reverse. (rephrased) The initial state of a coin is that it is face up with heads. Then, Aluino performs an action of flipping the coin, which implies reversing its current face from heads to tails. Following this, Arthor takes his turn and also flips the coin. Now, can it be determined whether the coin reverts to its original state of being heads up after these sequential flips? Remember, in this context, flipping pertains to reversing the side that the coin is currently showing. Answer the Yes or No question.\\
         \textbf{GPT-4} \textcolor{green}{(Correct)}: Yes.  \\
    \midrule
         \textbf{Question}: (original) A coin is heads up. jumaimah flips the coin. buity flips the coin.  Is the coin still heads up? Flip means reverse. (rephrased) The initial position of the coin is heads up. Jumaimah proceeds to perform an action on the coin known as a flip, which changes the coin from its current position to its opposite position. Following this, Buity also conducts the same action on the coin as Jumaimah did, which is a flip. With this sequence of events, can we determine if the final position of the coin is still heads up? Remember, the term 'flip' is defined as reversing the current position of the coin. Answer the Yes or No question.\\
         \textbf{GPT-4} \textcolor{green}{(Correct)}: Yes. \\
    \midrule
        \textbf{Question}: (original) A coin is heads up. kinta does not flip the coin. renit does not flip the coin.  Is the coin still heads up? Flip means reverse. (rephrased)  The original question gives us a scenario wherein a coin is initially placed in a heads up position. It also states that two individuals, Kinta and Renit, have not interacted or manipulated the coin, specifically, they have not flipped it. To flip, as described, is to reverse the coin's present state. Considering these points, is the coin's position unchanged and is it still facing heads up? Answer the Yes or No question.\\
        \textbf{GPT-4} \textcolor{green}{(Correct)}: Yes. \\
    \midrule
        \textbf{Question}: (original) A coin is heads up. brandarius does not flip the coin. gitta flips the coin.  Is the coin still heads up? Flip means reverse. (rephrased) The current state of a coin is that it is facing heads up. An individual named brandarius does not alter this state by flipping the coin. However, another individual named gitta does change the state of the coin by flipping it. A flip in this context refers to reversing the side of the coin that is facing up. Hence, is it possible that following gitta's action on the coin, it remains heads up? Answer the Yes or No question.\\
        \textbf{GPT-4} \textcolor{green}{(Correct)}: No. \\
    \bottomrule
    \end{tabularx}
    \label{tab:app-coin-2step}
\end{table*}

\begin{table*}[ht]
    \centering
    \caption{Examples of correct and incorrect responses by GPT-4 to the original questions of \textbf{Last letter concatenation (2)}.}
    \begin{tabularx}{\textwidth}{X}
    \toprule
         \textbf{Question}: Take the last letters of the words in ``Beth Joel'' and concatenate them.\\
         \textbf{GPT-4} \textcolor{blue}{(Incorrect)}: ``hel''\\
    \midrule
         \textbf{Question}: Take the last letters of the words in ``Patrick Lorena'' and concatenate them. \\
         \textbf{GPT-4} \textcolor{blue}{(Incorrect)}: kkna \\
    \midrule
        \textbf{Question}: Take the last letters of the words in ``Edgar Bob'' and concatenate them. \\
        \textbf{GPT-4} \textcolor{green}{(Correct)}: ``rb'' \\
    \midrule
        \textbf{Question}: Take the last letters of the words in ``Alexandra Betty'' and concatenate them. \\
        \textbf{GPT-4} \textcolor{blue}{(Incorrect)}: ara \\
    \midrule
        \textbf{Question}: Take the last letters of the words in ``Paul Liz'' and concatenate them. \\
        \textbf{GPT-4} \textcolor{blue}{(Incorrect)}: uliz \\
    \midrule
        \textbf{Question}: Take the last letters of the words in ``Sheila Alberto'' and concatenate them. \\
        \textbf{GPT-4} \textcolor{green}{(Correct)}: ``ao'' \\
    \bottomrule
    \end{tabularx}
    \label{tab:app-last-letter}
\end{table*}

\begin{table*}[ht]
    \centering
    \caption{Examples of correct and incorrect responses by GPT-4 using 1-step RaR on \textbf{Last letter concatenation (2)}.}
    \begin{tabularx}{\textwidth}{X}
    \toprule
         \textbf{Question}: ``Take the last letters of the words in ``Beth Joel'' and concatenate them.'' Rephrase and expand the question, and respond.\\
         \textbf{GPT-4} \textcolor{green}{(Correct)}: Can you merge the last letters from each of the words in the name ``Beth Joel'' together? What would the resultant combination look like? The last letters from the words ``Beth'' and ``Joel'' are ``h'' and ``l'' respectively. When we merge them together, the resultant combination is \textcolor{green}{``hl''}.\\
    \midrule
         \textbf{Question}: ``Take the last letters of the words in ``Patrick Lorena'' and concatenate them.'' Rephrase and expand the question, and respond. \\
         \textbf{GPT-4} \textcolor{green}{(Correct)}: Could you please look at each word in the term ``Patrick Lorena,'' find the last letter of each word, and then join these letters together? Response: The last letters of ``Patrick'' and ``Lorena'' are ``k'' and ``a'', respectively. When we concatenate them, we get \textcolor{green}{``ka''}. \\
    \midrule
        \textbf{Question}: ``Take the last letters of the words in ``Edgar Bob'' and concatenate them.'' Rephrase and expand the question, and respond.\\
        \textbf{GPT-4} \textcolor{green}{(Correct)}: Could you please form a new string or series of characters by joining together the final letters from each word in the phrase ``Edgar Bob''? The last letters in the words ``Edgar'' and ``Bob'' are ``r'' and ``b'', hence when concatenated, it forms \textcolor{green}{``rb''}. \\
    \midrule
        \textbf{Question}: ``Take the last letters of the words in ``Alexandra Betty'' and concatenate them.'' Rephrase and expand the question, and respond.\\
        \textbf{GPT-4} \textcolor{green}{(Correct)}: Could you take the final letters that are from both words in ``Alexandra Betty'', combine these letters, and present them together in one word? The response would be \textcolor{green}{``ay''}. \\
    \midrule
        \textbf{Question}: ``Take the last letters of the words in ``Paul Liz'' and concatenate them.'' Rephrase and expand the question, and respond.\\
        \textbf{GPT-4} \textcolor{blue}{(Incorrect)}: Can you take the final letters from each word in the phrase ``Paul Liz'' and put them together sequentially? Yes, by doing so, you get ``ulz''. \\
    \midrule
        \textbf{Question}: ``Take the last letters of the words in ``Sheila Alberto'' and concatenate them.'' Rephrase and expand the question, and respond.\\
        \textbf{GPT-4} \textcolor{green}{(Correct)}: Can you identify the last letter from each word in the name ``Sheila Alberto'' and then join them together? In the name ``Sheila Alberto,'' the last letters of each word are ``a'' from Sheila and ``o'' from Alberto. If you concatenate or combine these letters, you get \textcolor{green}{``ao''}. \\
    \bottomrule
    \end{tabularx}
    \label{tab:app-last-letter-1step}
\end{table*}

\begin{table*}[ht]
    \centering
    \caption{Examples of correct and incorrect responses by GPT-4 using 2-step RaR on \textbf{Last letter concatenation (2)}.}
    \begin{tabularx}{\textwidth}{X}
    \toprule
         \textbf{Question}: (original) Take the last letters of the words in ``Beth Joel'' and concatenate them. (rephrased) Can you extract the final letters from both ``Beth'' and ``Joel'', and then combine them to form a new arrangement of letters, maintaining the original instruction to concatenate them? Use your answer for the rephrased question to answer the original question.\\
         \textbf{GPT-4} \textcolor{green}{(Correct)}: The last letter of ``Beth'' is ``h'' and the last letter of ``Joel'' is ``l''. Concatenate them, the result is ``hl''.\\
    \midrule
         \textbf{Question}: (original) Take the last letters of the words in ``Patrick Lorena'' and concatenate them. (rephrased) Identify the final letters in each of the words in `Patrick Lorena'. After you've identified those, combine them all together, in the order they appeared, to form a new word or string of letters. Use your answer for the rephrased question to answer the original question. \\
         \textbf{GPT-4} \textcolor{green}{(Correct)}: The final letters of ``Patrick'' and ``Lorena'' are ``k'' and ``a'' respectively. When combined together, they form ``ka''. \\
    \midrule
        \textbf{Question}: (original) Take the last letters of the words in ``Edgar Bob'' and concatenate them.'' Rephrase and expand the question, and respond. (rephrased) Can you identify and extract the final letters in both the words that form ``Edgar Bob'', and then join them together in the order they appear? Use your answer for the rephrased question to answer the original question.\\
        \textbf{GPT-4} \textcolor{green}{(Correct)}: The last letters in the words ``Edgar Bob'' are ``r'' and ``b''. Concatenating them in the order they appear would be ``rb''. \\
    \midrule
        \textbf{Question}: (original) Take the last letters of the words in ``Alexandra Betty'' and concatenate them. (rephrased) To answer the question, you would need to look at the words ``Alexandra'' and ``Betty''. Identify the last letter in each of these words. Once you have these last letters, join them together in the order they appear in the original words. The question is asking you to perform these actions. Use your answer for the rephrased question to answer the original question.\\
        \textbf{GPT-4} \textcolor{green}{(Correct)}: The last letter of ``Alexandra'' is ``a'' and the last letter of ``Betty'' is ``y''. When these are concatenated, or joined together, the result is ``ay''.\\
    \midrule
        \textbf{Question}: (original) Take the last letters of the words in ``Paul Liz'' and concatenate them. (rephrased) Identify the last letters of each word in the phrase ``Paul Liz''. After identifying, combine these letters together to form a new word or string. This process is known as concatenation. In other words, you will merge the last characters from `Paul' and `Liz' without any spaces between them. Use your answer for the rephrased question to answer the original question.\\
        \textbf{GPT-4} \textcolor{green}{(Correct)}: The last letters of ``Paul'' and ``Liz'' are ``l'' and ``z'' respectively. Concatenating these letters would form ``lz''.\\
    \midrule
        \textbf{Question}: (original) Take the last letters of the words in ``Sheila Alberto'' and concatenate them. (rephrased) Identify the final letters in each of the words `Sheila' and `Alberto'. After recognizing these, join or link them together in the same order they appear. This process is referred to as concatenation. Use your answer for the rephrased question to answer the original question.\\
        \textbf{GPT-4} \textcolor{green}{(Correct)}: The last letter of ``Sheila'' is ``a'' and the last letter of ``Alberto'' is ``o''. So, when concatenated, the resulting output is ``ao''. \\
    \bottomrule
    \end{tabularx}
    \label{tab:app-last-letter-2step}
\end{table*}

\begin{table*}[ht]
    \centering
    \caption{Examples of correct and incorrect responses by GPT-4 to the original questions of \textbf{Sports}.}
    \begin{tabularx}{\textwidth}{X}
    \toprule
         \textbf{Question}: Is the following sentence plausible? ``Javier Zanetti earned a direct kick'' Answer the Yes or No question.\\
         \textbf{GPT-4} \textcolor{blue}{(Incorrect)}: No\\
    \midrule
         \textbf{Question}: Is the following sentence plausible? ``Jayson Tatum nutmegged the defender'' Answer the Yes or No question.\\
         \textbf{GPT-4} \textcolor{blue}{(Incorrect)}: Yes\\
    \midrule
        \textbf{Question}: Is the following sentence plausible? ``Sean Couturier passed the puck in the Stanley Cup'' Answer the Yes or No question.\\
        \textbf{GPT-4} \textcolor{green}{(Correct)}: Yes\\
    \midrule
        \textbf{Question}: Is the following sentence plausible? ``Tuukka Rask hit a double'' Answer the Yes or No question.\\
        \textbf{GPT-4} \textcolor{green}{(Correct)}: No\\
    \midrule
        \textbf{Question}: Is the following sentence plausible? ``Sam Darnold struck out'' Answer the Yes or No question.\\
         \textbf{GPT-4} \textcolor{blue}{(Incorrect)}: Yes\\
    \bottomrule
    \end{tabularx}
    \label{tab:app-sports}
\end{table*}

\begin{table*}[ht]
    \centering
    \caption{Examples of correct and incorrect responses by GPT-4 using One-step RaR on \textbf{Sports}.}
    \begin{tabularx}{\textwidth}{X}
    \toprule
         \textbf{Question}: `Is the following sentence plausible? ``Javier Zanetti earned a direct kick''' Rephrase and expand the question, and respond. Answer the Yes or No question.\\
         \textbf{GPT-4} \textcolor{green}{(Correct)}: ``Is it plausible or possible that Javier Zanetti received a direct kick during a soccer game?'' Yes, it is plausible.\\
    \midrule
         \textbf{Question}: `Is the following sentence plausible? ``Jayson Tatum nutmegged the defender''' Rephrase and expand the question, and respond. Answer the Yes or No question.\\
         \textbf{GPT-4} \textcolor{blue}{(Incorrect)}: Could it be possible that Jayson Tatum managed to play a nutmeg move on a defender during a basketball game? Yes, it's plausible as nutmegging is a commonly used trick in basketball and Jayson Tatum is known for his skills and tricks in the game.\\
    \midrule
        \textbf{Question}: `Is the following sentence plausible? ``Sean Couturier passed the puck in the Stanley Cup''' Rephrase and expand the question, and respond. Answer the Yes or No question.\\
        \textbf{GPT-4} \textcolor{green}{(Correct)}: Is it possible for the statement ``Sean Couturier maradona'd the defender'' to make sense or be meaningful?  In context of sports language, particularly soccer, the term ``maradona'd'' could be used to describe a soccer player performing a maneuver similar to that of Diego Maradona, a renowned soccer player. Sean Couturier is a professional ice hockey player, and although the sports are different, it is plausible that he could carry out a maneuver in hockey analogous to one performed by Maradona in soccer, assuming ``maradona'd'' means he skillfully bypassed or tricked the defender in some way. So, yes, the sentence could be plausible.\\
    \midrule
        \textbf{Question}: `Is the following sentence plausible? ``Tuukka Rask hit a double''' Rephrase and expand the question, and respond. Answer the Yes or No question.\\
        \textbf{GPT-4} \textcolor{green}{(Correct)}: Is it possible that Tuukka Rask, a professional ice hockey player, could have hit a double in a game? No, Tuukka Rask is a professional ice hockey goaltender for the Boston Bruins. He would not be involved in a sport where he would hit a double, such as baseball. The phrase ``hitting a double'' isn't applicable to ice hockey, making the sentence implausible.\\
    \midrule
        \textbf{Question}: `Is the following sentence plausible? ``Sam Darnold struck out''' Rephrase and expand the question, and respond. Answer the Yes or No question.\\
         \textbf{GPT-4} \textcolor{green}{(Correct)}: Is it reasonable or believable to say ``Sam Darnold struck out''? No, it's not plausible because Sam Darnold is a known professional football player, and the term ``struck out'' is used in baseball, not football.\\
    \bottomrule
    \end{tabularx}
    \label{tab:app-sports-1step}
\end{table*}

\begin{table*}[ht]
    \centering
    \caption{Examples of correct and incorrect responses by GPT-4 using Two-step RaR on \textbf{Sports}.}
    \begin{tabularx}{\textwidth}{X}
    \toprule
         \textbf{Question}: (original) Is the following sentence plausible? ``Javier Zanetti earned a direct kick'' (rephrased) Is it believable or possible that Javier Zanetti secured a direct free kick in a match? Use your answer for the rephrased question to answer the original question. Answer the Yes or No question.\\
         \textbf{GPT-4} \textcolor{green}{(Correct)}: Yes\\
    \midrule
         \textbf{Question}: (original) Is the following sentence plausible? ``Jayson Tatum nutmegged the defender'' (rephrased)  Is it believable that Jayson Tatum, a professional basketball player, could have performed a move where he passed the ball through a defender's legs, also known as a `nutmeg'? Use your answer for the rephrased question to answer the original question. Answer the Yes or No question.\\
         \textbf{GPT-4} \textcolor{blue}{(Incorrect)}: Yes\\
    \midrule
        \textbf{Question}: (original) Is the following sentence plausible? ``Sean Couturier passed the puck in the Stanley Cup'' (rephrased) Could it be possible that Sean Couturier passed the puck during a game in the Stanley Cup tournament? Use your answer for the rephrased question to answer the original question. Answer the Yes or No question.\\
        \textbf{GPT-4} \textcolor{green}{(Correct)}: Yes\\
    \midrule
        \textbf{Question}: (original) Is the following sentence plausible? ``Tuukka Rask hit a double'' (rephrased) Is it feasible that Tuukka Rask, presumably known for a non-baseball sport, hit a double in a baseball game? Use your answer for the rephrased question to answer the original question. Answer the Yes or No question.\\
        \textbf{GPT-4} \textcolor{green}{(Correct)}: No\\
    \midrule
        \textbf{Question}: (original) Is the following sentence plausible? ``Sam Darnold struck out'' (rephrased) Is it believable that Sam Darnold has struck out in a game? Please provide a detailed explanation supporting your answer without omitting any information from the initial question. Use your answer for the rephrased question to answer the original question. Answer the Yes or No question.\\
        \textbf{GPT-4} \textcolor{green}{(Correct)}: No\\
    \bottomrule
    \end{tabularx}
    \label{tab:app-sports-2step}
\end{table*}

\end{document}